\pdfoutput=1
\documentclass[11pt]{article}
\usepackage{arxiv}

\usepackage{times}
\usepackage[T1]{fontenc}
\usepackage[utf8]{inputenc}
\usepackage{graphicx}
\usepackage{multirow}
\usepackage{amsmath,amssymb,amsfonts}
\usepackage[title]{appendix}
\usepackage{xcolor}
\usepackage[mathscr]{euscript}
\usepackage{textcomp}
\usepackage{manyfoot}
\usepackage{booktabs}
\usepackage{algorithm}
\usepackage{algorithmicx}
\usepackage{algpseudocode}
\usepackage{listings}
\usepackage[numbers]{natbib}
\usepackage[colorlinks]{hyperref}
\hypersetup{
pdftitle={TF1-EN-3M: Three Million Synthetic Moral Fables from Open Language Models},
pdfauthor={Mihai Nadas, Laura Diosan, Andreea Tomescu, Andrei Piscoran},
pdfkeywords={synthetic data generation, fable generation, moral reasoning, language models, dataset curation},
bookmarksnumbered,
pdfstartview={FitH},
colorlinks=true,
linkcolor=blue,
citecolor=blue,
urlcolor=blue,
}

\begin{document}
\title{TF1-EN-3M: Three Million Synthetic Moral Fables from Open Language Models}

\author{
  Mihai Nadăș\textsuperscript{1} \and
  Laura Dioșan\textsuperscript{1} \and
  Andreea Tomescu\textsuperscript{2} \and
  Andrei Pișcoran\textsuperscript{2} \\[1ex]
  \textsuperscript{1}Babeș-Bolyai University, Cluj-Napoca, Romania \\
  \textsuperscript{2}KlusAI Labs, Cluj-Napoca, Romania \\
  \texttt{mihai.nadas@ubbcluj.ro}
}

\maketitle

\begin{abstract}
Moral stories are a time-tested vehicle for transmitting values, yet modern NLP lacks a large, \emph{structured} corpus that couples coherent narratives with explicit ethical lessons. We present \textsc{TF1-EN-3M}, to our knowledge the first open dataset of \textbf{three million} English-language fables generated exclusively by instruction-tuned models no larger than 8\,B parameters. Each story follows a six-slot scaffold (character $\rightarrow$ trait $\rightarrow$ setting $\rightarrow$ conflict $\rightarrow$ resolution $\rightarrow$ moral), produced through a combinatorial prompt engine that guarantees genre fidelity while covering a broad thematic space.

A fully reproducible evaluation pipeline employs a panel of open-weight LLM judges from distinct model families, scoring grammar, creativity, moral clarity, and template adherence, complemented by reference-free diversity and readability metrics. Among ten open-weight generator candidates, an 8\,B-parameter Llama-3 variant delivers the best quality--cost trade-off, producing high-scoring fables on consumer hardware at ${\approx}\$0.135$ per 1\,000 fables.

We release the dataset, generation code, evaluation scripts, and full metadata under a permissive license, enabling exact reproducibility and cost benchmarking. \textsc{TF1-EN-3M} opens avenues for research in instruction following, narrative intelligence, value alignment, and child-friendly educational AI---demonstrating that large-scale moral storytelling requires neither proprietary giant models nor proprietary evaluation infrastructure.
\end{abstract}

\keywords{synthetic data generation \and fable generation \and moral reasoning \and language models \and dataset curation}
\section{Introduction}
\label{sec:introduction}

Stories that impart moral lessons --- \textbf{fables} --- have long served as a compelling medium for teaching values and social norms. As Guan \emph{et al.}\ \citep{guan_corpus_2022} note, \emph{"Teaching morals is one of the most important purposes of storytelling"}. Traditionally, fables feature anthropomorphic characters and conclude with explicit morals that connect concrete events to abstract ethical principles. However, classical collections, such as Aesop’s Fables \citep{aesop_aesops_2002}, are limited in size, which restricts data-driven approaches to modeling moral storytelling.

In recent years, advances in large language models (LLMs) have unlocked new possibilities for \textbf{synthetic data generation} in natural language processing. Researchers have increasingly explored LLMs as a cost-effective alternative to manual annotation, generating high-quality datasets that span various domains — from classification to dialogue \citep{long_llms-driven_2024}. Projects like \textbf{TinyStories} \citep{eldan_tinystories_2023} have demonstrated that even models with fewer than 10M parameters can learn to produce coherent narratives when trained on carefully curated synthetic data \citep{gunasekar_textbooks_2023}. This progress underscores the potential for LLM-generated corpora to foster development in the realm of \textbf{open model} research, especially in low-data regimes. A recent comprehensive survey further contextualizes these advances across both text and code domains, highlighting key methods such as prompt-based generation, retrieval-augmented pipelines, and reinforcement-driven refinement \citep{nadas_synthetic_2025}.

\subsection{Research Questions}
Motivated by (i) the absence of any large-scale, structured fable corpus for data‐driven moral storytelling and (ii) the need to understand whether compact, open‐weight models can reliably generate such narratives, we investigate:
\begin{itemize}
    \item \textbf{RQ1:} How effective is a combinatorial prompt expansion methodology in generating diverse and high-quality fables using LLMs?
    \item \textbf{RQ2:} Which open-weight LLMs are best suited for fable generation under resource constraints?
\end{itemize}

Motivated by these questions and emerging trends, we set out to construct a \textbf{large-scale dataset of fables} using modern open-weight LLMs. Our objective is to capture narratives with clear pedagogical intent and a consistent structure, suitable for applications in \textbf{story generation}, \textbf{moral reasoning}, and \textbf{instruction following}. Key challenges include: (a) systematically covering a wide \textbf{diversity of scenarios} via prompt design, and (b) ensuring the \textbf{quality and consistency} of generated stories while relying on \textbf{resource-limited models} that run on consumer-grade hardware rather than on expensive API-based systems.

Our approach employs a \textbf{structured prompt template} that encodes classic fable elements—protagonist, character trait, setting, conflict, resolution, and moral. By combinatorially expanding lists for each element, we generate millions of unique prompts that span an expansive range of moral narratives.

To determine the most suitable model for full-scale generation, we conduct an \textbf{extensive evaluation of candidate models} (including variants from Llama-3, Mistral, DeepSeek, and others). Evaluation proceeds along two complementary axes:

\begin{enumerate}
    \item A \textbf{multi-judge LLM panel} comprising open-weight models from distinct families, each scoring grammar, creativity, moral clarity, and prompt adherence---following best practices for reproducible LLM-based evaluation \citep{liu_g-eval_2023, fu_gptscore_2023}.
    
    \item A set of \textbf{reference-free diversity and readability metrics}, including \emph{Self-BLEU} \citep{zhu_texygen_2018}, \emph{Distinct-n} \citep{li_diversity-promoting_2016}, and the \emph{Flesch Reading Ease} score \citep{kincaid_derivation_1975}, which provide an independent signal of lexical variety, textual novelty, and accessibility.
\end{enumerate}

This multi-perspective assessment ultimately leads us to choose \textbf{Llama-3.1-8B-Instruct} as the model for generating the TF1-EN-3M dataset. While other models achieved marginally higher scores in some dimensions, Llama-3.1-8B consistently balanced narrative quality with stylistic simplicity, producing fables best aligned with the 4--7 age group, as the naturally limited lexical range of this audience makes the corpus especially conducive to parameter-efficient fine-tuning of small downstream models.

\subsection{Contributions}
Our work makes the following key contributions:
\begin{enumerate}
    \item \textbf{TF1-EN-3M Dataset:} We release, to our knowledge, the \textbf{first large-scale open dataset of synthetic fables in English} (3,000,000 stories), providing a novel resource for research in story generation, moral reasoning, and instruction following.
    \item \textbf{Efficient Open-Weight Generation:} We demonstrate that \textbf{high-quality, instructive content} can be generated at scale using \textbf{relatively small open models} (1--8B parameters) deployable on consumer-grade hardware, at a total cost of \$405 for the entire corpus \citep{eldan_tinystories_2023}.
    \item \textbf{Reproducible Multi-Judge Evaluation:} We propose a fully reproducible evaluation framework using a panel of \textbf{open-weight LLM judges} from distinct model families, combined with reference-free diversity and readability metrics, enabling transparent multi-criteria model selection without reliance on proprietary evaluation infrastructure.
\end{enumerate}

The TF1-EN-3M dataset is published on Hugging Face Hub under the identifier \textbf{\texttt{klusai/ds-tf1-en-3m}} \citep{noauthor_klusaids-tf1-en-3m_nodate}, and the full code to regenerate and evaluate it is available in the TinyFabulist repository.

The remainder of this paper is organized as follows. Section~\ref{sec:related-work} reviews related work across synthetic data generation, moral NLP, LLM-based evaluation, and template-driven generation. Section~\ref{sec:prompt-design} formalizes our template-driven prompt schema and the large-scale dataset generation pipeline. Section~\ref{sec:eval-llms} presents our evaluation framework and compares the performance of ten open-weight models. In Section~\ref{sec:dataset} we describe the TF1-EN-3M dataset---its structure, metadata schema, generation costs, and public release---and in Section~\ref{sec:discussion} we reflect on our results, explore practical applications, and outline threats to validity. Finally, Section~\ref{sec:conclusion} summarizes our contributions and suggests directions for future work.

\section{Related Work}
\label{sec:related-work}

Our work draws on four active research threads: synthetic data generation at scale, moral and narrative NLP, LLM-based evaluation, and template-driven story generation.

\subsection{Synthetic Data Generation}
The use of LLMs to generate large-scale training corpora has gained substantial momentum. \textbf{TinyStories} \citep{eldan_tinystories_2023} demonstrated that GPT-3.5/4 can produce millions of children's stories suitable for training sub-10M-parameter models. The ``textbooks are all you need'' paradigm \citep{gunasekar_textbooks_2023} further showed that curated synthetic data can match or exceed the utility of larger web-crawled corpora. Self-Instruct \citep{wang_self-instruct_2023} bootstraps fine-tuning data from model outputs, while Persona Hub \citep{ge_scaling_2024} scales instruction data via billions of persona templates. Scaling laws for synthetic data \citep{muennighoff_scaling_2023} provide theoretical grounding for these approaches. A comprehensive survey contextualizes advances across text and code domains \citep{nadas_synthetic_2025}. Our work follows this paradigm but targets the moral-fable genre specifically, using combinatorial slot-filling rather than persona- or QA-based prompts.

\subsection{Moral and Narrative NLP}
Story generation has evolved from heuristic planners with fixed plot templates \citep{riedl_narrative_2010} to neural approaches using hierarchical generation \citep{fan_hierarchical_2018} and plan-and-write frameworks \citep{yao_plan-and-write_2019}. In the moral reasoning domain, STORAL \citep{guan_corpus_2022} assembled human-authored moral tales and showed that off-the-shelf models struggle to align plot events with morals, motivating the need for specialised training data. The Story Cloze Test \citep{mostafazadeh_corpus_2016} evaluates narrative comprehension. Our dataset complements these resources by offering a synthetic, massively parallel corpus of moral fables with explicit ethical lessons, suitable for both fine-tuning narrative generators and probing moral-reasoning capabilities.

\subsection{LLM-Based Evaluation}
Evaluating open-ended narratives with traditional reference-based metrics (BLEU, ROUGE) often fails to capture creativity and thematic coherence \citep{nguyen_comparative_2024}. Recent frameworks use LLMs as learned evaluators: G-Eval \citep{liu_g-eval_2023} prompts GPT-4 with explicit rubrics and chain-of-thought to produce human-aligned scores; GPTScore \citep{fu_gptscore_2023} similarly leverages LLM judgments across multiple axes. However, relying on a single proprietary judge introduces reproducibility concerns and systematic bias \citep{zheng_judging_2023}. Multi-agent evaluation frameworks have been shown to outperform single-judge setups through debate and meta-judging protocols \citep{chan2023chateval, xu2025metajudge}. Self-preference bias---where LLM judges favour outputs from their own family---has been documented across multiple studies \citep{wataoka2024selfpref, pan2024selfpref, chen2025selfpref}, motivating family-diverse evaluation panels. Chain-of-thought and reasoning-capable judges have been shown to improve judgment quality, particularly for nuanced assessments \citep{wei2022chainofthought}. Recent surveys of LLM-as-judge methodology provide further methodological grounding, covering bias taxonomies, calibration strategies, and best practices for multi-judge evaluation \citep{gu2024surveyllmjudge}.

\subsection{Template-Driven Generation}
Template-based prompt design has shown effectiveness in guiding language models toward structured and purposeful text generation \citep{liu_what_2021, mishra_reframing_2022}. Instruction-based controllability in open-ended generation \citep{honovich_unnatural_2023} demonstrates that explicit structural and stylistic constraints can be combined with creative freedom. Our six-slot scaffold (character, trait, setting, conflict, resolution, moral) builds on these insights, providing both genre fidelity and broad thematic coverage through combinatorial expansion.

\section{Prompt design and dataset generation}
\label{sec:prompt-design}
We first define a structured prompt schema informed by narrative theory and prior template-based generation work \citep{liu_what_2021, mishra_reframing_2022}, then construct a controlled value space from which to sample inputs. Finally, we specify sampling strategies and length constraints to balance diversity, coherence, and computational tractability.

\subsection{Prompt Schema and Value Space}
Our goal is to generate texts that are \textbf{diverse}, \textbf{coherent}, and \textbf{tailored to the fable form}. To achieve consistency in structure while covering a broad range of content, we employ a \textbf{template-driven prompt design}, which has shown effectiveness in guiding language models towards structured and purposeful text generation \citep{liu_what_2021, mishra_reframing_2022}. Specifically, we designed our prompt templates around six key \textbf{fable elements} identified through analysis of traditional narrative structures:

\begin{itemize}
\item \textbf{Character:} The protagonist, frequently depicted as an animal or human archetype (e.g., \emph{fox}, \emph{woodcutter}).
\item \textbf{Trait:} A notable attribute that shapes character behavior and story outcomes (e.g., \emph{greedy}, \emph{kind}, \emph{lazy}).
\item \textbf{Setting:} Contextual locations that ground narratives within recognizable environments (e.g., \emph{in a dense forest}, \emph{on a farm}, \emph{in a bustling town}).
\item \textbf{Conflict:} Challenges or dilemmas central to narrative tension (e.g., \emph{loses their food to someone's trick}, \emph{must choose between truth and lies}).
\item \textbf{Resolution:} Methods by which conflicts resolve, often demonstrating moral or ethical judgments (e.g., \emph{the character learns sharing}, \emph{the trickster is exposed}).
\item \textbf{Moral:} Explicit lessons underscoring narrative intent (e.g., \emph{"Don't cry wolf unless you mean it"}, \emph{"Honesty is the best policy"}).
\end{itemize}

\paragraph{Prompt Construction:} We developed a prompt template incorporating these elements into a structured narrative prompt. The general form reads: \begin{quote}\small
Create a fable based on the following elements. Weave them naturally into a story:\\
-- Main Character: A [Trait] [Character]\\
-- Setting: [Setting]\\
-- Challenge: [Conflict]\\
-- Outcome: [Resolution]\\
-- Teaching: [Moral]\\[.5ex]
\end{quote}
This explicit structuring ensures that generated stories follow conventional narrative patterns, providing clear beginnings, middles, and ends, alongside an explicit moral lesson, thus directly addressing the documented challenge of aligning generated stories with specified outcomes \citep{fan_hierarchical_2018, yao_plan-and-write_2019}.

Beyond slot-filling, we augment our prompts with stylistic constraints to promote narrative quality and adherence to the fable genre. The template includes explicit instructions for how the model should realize each element within the story: to begin with vivid scene-setting, avoid naming characters (instead referring to their role and trait), use meaningful dialogue, show the character’s growth implicitly (rather than stating it), and conclude with a clear, moral-driven resolution. These stylistic cues serve as soft controls, guiding the language model toward coherent and pedagogically meaningful outputs while maintaining creative freedom. This approach draws on narrative writing principles such as “show, don’t tell” and mimics the tone and form of traditional fables, which often rely on concise, evocative storytelling to convey timeless ethical lessons. Similar strategies have been shown to improve narrative coherence and genre alignment in neural story generation \citep{fan_hierarchical_2018, yao_plan-and-write_2019}, and are consistent with recent work on instruction-based controllability in open-ended generation tasks \citep{honovich_unnatural_2023, mishra_reframing_2022}.

\begin{figure}[!htbp]
  \centering
  \includegraphics[width=1\linewidth]{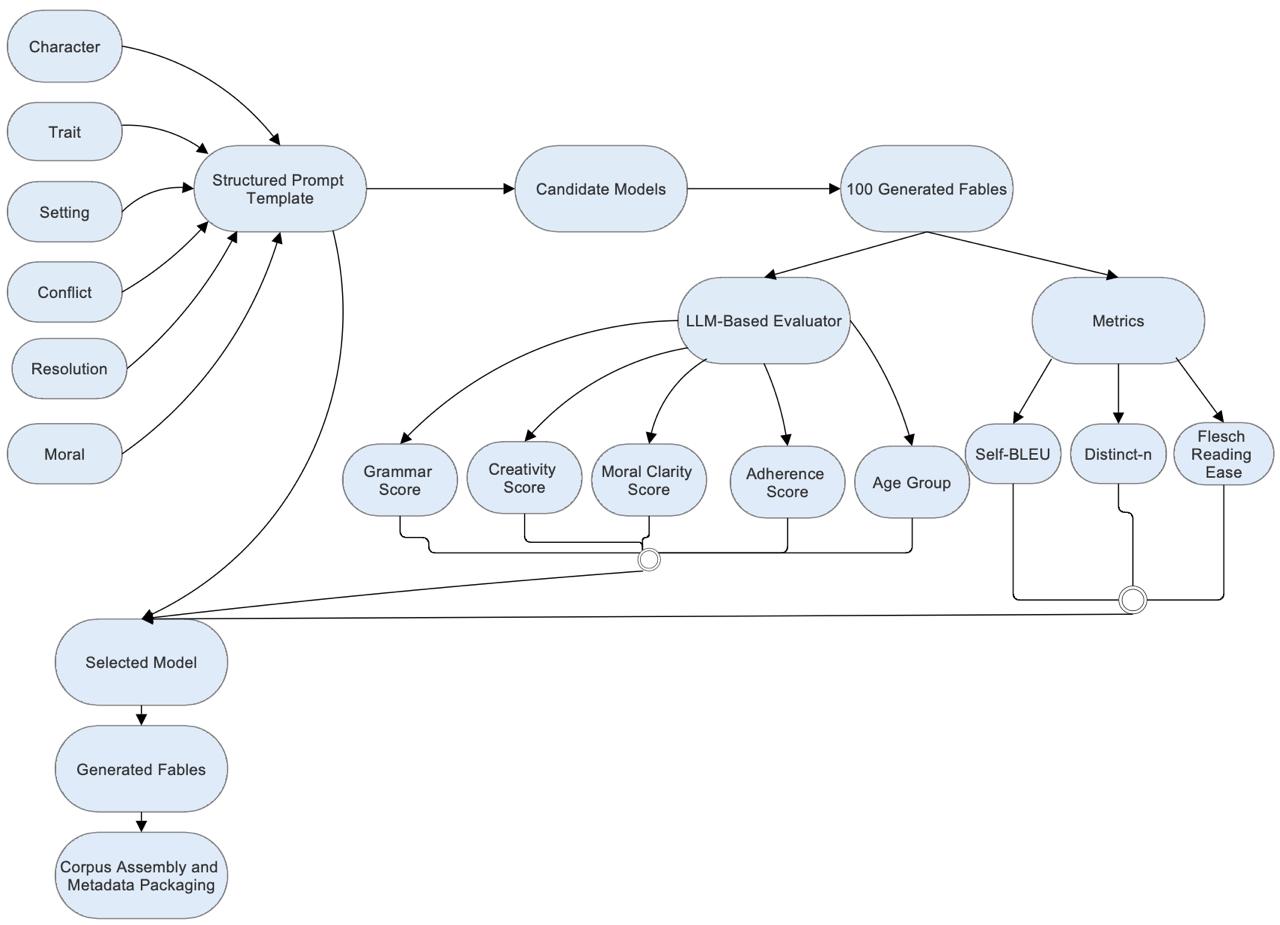}
  \caption{Full pipeline for generating TF1-EN-3M.}
  \label{fig:full_pipeline}
\end{figure}

\paragraph{System Message Design: }

In addition to the structured narrative template, we provided each language model with a dedicated system message that framed the task as moral fable writing. This instruction established expectations for output quality, genre-specific constraints, and age-appropriate narrative style.

The system message emphasized three key principles:
\begin{itemize}
    \item \textbf{Imaginative and coherent storytelling,}
    \item \textbf{Audience awareness}, including suitability for young readers,
    \item \textbf{Adherence to the classic fable format}, composed of six core elements: character, trait, setting, conflict, resolution, and moral.
\end{itemize}

It also defined five distinct age groups (A–E) used later during evaluation to assess target audience alignment. These audience categories helped our LLM-based critic assign each story to an appropriate demographic bracket.

The system message complements the structured prompt by establishing tone, ethical clarity, and pedagogical intent — critical elements in moral storytelling. It acts as a global instruction layer, ensuring genre fidelity across millions of independently sampled and generated fables.

\paragraph{Value Space and Prompt Design Variables}

We curated structured value sets for each slot in the prompt template, forming a controlled input space for synthetic generation. These sets were derived from a combination of classical fables, common moral themes, and creative extensions, ensuring both cultural relevance and narrative diversity, and were manually populated by our research team.

We abstract the input space by introducing six formal parameters:
\[
n,\;m,\;k,\;c,\;r,\;l
\]
where
\begin{itemize}
  \item $n$ = \# of Character options,
  \item $m$ = \# of Trait options,
  \item $k$ = \# of Setting options,
  \item $c$ = \# of Conflict options,
  \item $r$ = \# of Resolution options,
  \item $l$ = \# of Moral options.
\end{itemize}
The total combinatorial value space is thus
\[
T \;=\; n \times m \times k \times c \times r \times l.
\]
By parametrizing in this way, we cleanly separate the schema definition from the concrete instantiation. Concrete parameter values are specified in Section~\ref{sec:eval-llms}.

\subsection{Full System Message and Prompt Template}
Below we present the exact system message and prompt template that drive the generator.\\
\\
\textbf{System Message}
\begin{verbatim}
You are a world-class creative assistant that generates captivating and morally-driven fables based 
on structured inputs.
Each fable must be:
- Imaginative and coherent.
- Appropriate for a wide audience, including young readers.
- Structured around a classic fable format (character, setting, conflict, resolution, and moral).

Age groups are defined as:
- A: 3 years or under
- B: 4-7 years
- C: 8-11 years
- D: 12-15 years
- E: 16 years or above
\end{verbatim}
\textbf{Prompt Template}
\begin{verbatim}
Create a fable based on the following elements. Weave them naturally into a story:
- Main Character: a {{trait}} {{character}}
- Setting: a {{setting}} where our story unfolds
- Challenge: {{conflict}}
- Outcome: {{resolution}}
- Teaching: {{moral}}
The fable should:
- Be appropriate for age group B (4-7 years)
- Use simple vocabulary that 4-7 year olds can understand
- Use concrete rather than abstract language
- Begin with vivid scene-setting
- Not use names for the characters, instead use the trait and character
- Include meaningful but simple dialogue
- Show (don't tell) the character's growth
- End with a clear connection to the moral
Keep the story concise but engaging, around 250 words.
\end{verbatim}

\subsection{Evaluation Methodology}
\label{sec:evaluation-method}

Evaluating open-ended narrative generation is notoriously challenging. Recent studies have explored the use of large language models (LLMs) as learned evaluators, prompting them with rubric-driven instructions to produce multi-axis scores that better align with human preferences \citep{shen_large_2023, liu_g-eval_2023}.

\paragraph{Multi-Judge LLM Panel}
Building on the G-Eval framework \citep{liu_g-eval_2023} and similar work in summarization and dialogue evaluation \citep{shen_large_2023}, we assemble a panel of \textbf{three open-weight LLM judges} from distinct model families. Each judge receives the same rubric-driven prompt and assigns each fable a score from 1 to 10 along four dimensions:
\begin{itemize}
  \item \textbf{Grammar \& Style}: linguistic correctness and syntactic fluency,
  \item \textbf{Creativity}: narrative originality and inventiveness,
  \item \textbf{Moral Clarity}: explicitness and relevance of the ethical lesson,
  \item \textbf{Prompt Adherence}: fidelity to the template's structural and stylistic constraints.
\end{itemize}
The panel consists of: (1)~\textbf{Granite 4.1 30B} (IBM), a dense reasoning model; (2)~\textbf{EXAONE 3.5 32B} (LG AI Research), trained on a distinct data mix; and (3)~\textbf{Granite 3.3 8B} (IBM), serving as a lightweight arbiter. We note that two of three judges share the Granite lineage; however, Granite 3.3 and 4.1 differ substantially in architecture, training data, and generation (3.3 is an earlier, smaller model), and the panel's cross-vendor diversity is ensured by EXAONE. All three run locally on Apple M3 Ultra via Ollama at GGUF quantisation, ensuring full reproducibility without API costs. We additionally evaluate with \textbf{GPT-o4-mini} (via OpenRouter) as a proprietary reference point, enabling direct comparison of open-weight versus closed-source judge rankings. Using judges from multiple model families mitigates self-preference bias---where judges favour outputs from their own model family---a phenomenon documented in the LLM-as-judge literature \citep{zheng_judging_2023}. We report inter-judge agreement via weighted Cohen's kappa and verify ranking stability through a weight ablation study.

\paragraph{Age Group Classification}  
To ensure our fables are appropriate for different developmental stages, we also prompt the primary judge (Granite 4.1 30B) to classify each story into one of five age brackets—A (0–3 yrs), B (4–7 yrs), C (8–11 yrs), D (12–15 yrs), E (16+ yrs)—based on vocabulary simplicity, thematic maturity, and syntactic complexity. This mirrors methodologies in microfiction readability research \citep{eldan_tinystories_2023} and helps us verify that generated content aligns with intended educational use.

\paragraph{Reference-Free Metrics}  
Complementing these subjective evaluations, we compute three corpus-level, unreferenced statistics:
\begin{itemize}
  \item \textbf{Self-BLEU} \citep{zhu_texygen_2018}: quantifies intra-set redundancy by evaluating each story against the rest (lower values indicate greater diversity),
  \item \textbf{Distinct-n} \citep{li_diversity-promoting_2016}: measures the proportion of unique $n$-grams (we report $n=1$) as a proxy for lexical richness,
  \item \textbf{Flesch Reading Ease} \citep{kincaid_derivation_1975}: assesses overall readability via sentence and syllable counts, critical for age-appropriate comprehension.
\end{itemize}

By combining LLM-based critique, age-level classification, and automated diversity/readability metrics, our hybrid evaluation pipeline offers a robust, multi-faceted assessment of fable quality—addressing the known limitations of purely reference-based methods and reflecting best practices for generative evaluation \citep{fu_gptscore_2023}.

\section{LLM Evaluation and Model Selection}
\label{sec:eval-llms}

\subsection{Experimental Setup}
To explore how well our synthetic fables hold up under scrutiny, we first generated up to 
\(\texttt{MAX} = 3{,}000{,}000\) 
unique prompts from the full combinatorial space 
\[
T = n \times m \times k \times c \times r \times l,
\]
with each of the six parameters \(n,m,k,c,r,l\) set to 100.  We sampled uniformly at random, enforcing three gentle safeguards to preserve diversity and balance:
\begin{itemize}
  \item \emph{Uniqueness}—we discarded any duplicate prompts;
  \item \emph{Frequency filtering}—we downsampled overly common conflict–moral pairings;
  \item \emph{Coverage balancing}—we made sure each slot appeared roughly the same number of times.
\end{itemize}
Each fable was capped at max 1000 tokens, a length that aligns well with traditional fable structure \citep{naimou_short_2021}.  All used models were decoded using a consistent setup---temperature set to \(0.7\), with default top-p (i.e., 1.0), and greedy decoding disabled as a result of these parameters---to ensure a fair apples-to-apples comparison \citep{zhang_trading_2020}.

\subsection{Models Considered}
We evaluated ten instruction‐tuned, open‐weight LLMs deployable on consumer GPUs (<24 GB VRAM):
\begin{enumerate}
  \item \textbf{SmolLM2-1.7B-Instruct} \citep{noauthor_huggingfacetbsmollm2-17b-instruct_2025}
  \item \textbf{Aya-23-8B} \citep{noauthor_cohereforaiaya-23-8b_2025}
  \item \textbf{Llama-3.2-1B-Instruct} \citep{noauthor_meta-llamallama-32-1b-instruct_2024}
  \item \textbf{Llama-3.1-8B-Instruct} \citep{noauthor_meta-llamallama-31-8b-instruct_2024}
  \item \textbf{Llama-3.1-Tulu-3-8B} \citep{noauthor_meta-llamallama-31-8b-instruct_2024}
  \item \textbf{Mistral-7B-Instruct-v0.3} \citep{noauthor_mistralaimistral-7b-instruct-v03_nodate}
  \item \textbf{Qwen2.5-7B-Instruct} \citep{noauthor_qwenqwen25-7b-instruct_2025}
  \item \textbf{deepseek-llm-7b-chat} \citep{noauthor_deepseek-aideepseek-llm-7b-chat_2024}
  \item \textbf{Phi-3-mini-4k-instruct} \citep{noauthor_microsoftphi-3-mini-4k-instruct_2025}
  \item \textbf{Falcon3-7B-Instruct} \citep{noauthor_tiiuaefalcon3-7b-instruct_2025}
\end{enumerate}
For evaluation, each model generated fables from 100 randomly sampled prompts (1{,}000 fables total across ten models) under identical decoding hyperparameters.

\subsection{Results and Interpretation}

\paragraph{LLM‐based Evaluation}
Table~\ref{tab:llm_9col_updated} presents a head-to-head comparison of our models across Grammar, Creativity, Moral Clarity, and Prompt Adherence, along with the aggregate mean, token counts, and inference times.  

\begin{table*}[!htbp]
\centering
\caption{Multi-judge evaluation of fable generation (1--10 scale, averaged across three open-weight judges), plus generation metadata. Highest values in the first five metric columns (Grammar--Mean) are \textbf{bolded}; lowest latency is \textbf{bolded}. $\dagger$\,Token counts unavailable due to a logging issue in the Phi-3 inference endpoint.}
\label{tab:llm_9col_updated}
\resizebox{\textwidth}{!}{%
\renewcommand{\arraystretch}{1.2}
\begin{tabular}{|l|c|c|c|c|c|c|c|c|}
\hline
\textbf{Model} & 
\textbf{Grammar} & 
\textbf{Creativity} & 
\textbf{Moral Clarity} & 
\textbf{Adherence} & 
\textbf{Mean} & 
\textbf{Input Tokens} & 
\textbf{Output Tokens} & 
\textbf{Latency (s)} \\
\hline
Aya-23-8B                 & 8.36  & 6.64  & 8.03  & 8.35  & 7.85  & 171.8 & 500.6 & 257.89 \\
SmolLM2-1.7B-Instruct     & 8.17  & 6.27  & 7.75  & 7.89  & 7.52  & 174.3 & 414.7 & 17.58  \\
Qwen2.5-7B-Instruct       & 8.64  & 6.77  & 8.27  & 8.92  & 8.15  & 182.6 & 404.2 & 17.72  \\
Llama-3.1-Tulu-3-8B       & 8.77  & \textbf{7.09}  & \textbf{8.43}  & \textbf{9.24}  & \textbf{8.38}  & 181.5 & 368.5 & 16.91  \\
deepseek-llm-7b-chat      & 8.28  & 6.49  & 8.04  & 8.16  & 7.74  & 189.3 & 439.8 & 82.36  \\
Llama-3.1-8B-Instruct     & \textbf{8.75}  & 6.90  & 8.22  & 9.16  & 8.26  & 181.5 & 337.6 & 28.87  \\
Llama-3.2-1B-Instruct     & 8.36  & 6.37  & 7.68  & 8.08  & 7.62  & 181.5 & 358.5 & \textbf{16.69}  \\
Phi-3-mini-4k-instruct    & 8.57  & 6.80  & 8.23  & 8.86  & 8.11  &  N/A\textsuperscript{$\dagger$} &  N/A\textsuperscript{$\dagger$} & 40.76  \\
Mistral-7B-Instruct-v0.3  & 8.55  & 6.79  & 8.36  & 8.73  & 8.11  & 201.1 & 426.4 & 40.70  \\
Falcon3-7B-Instruct       & 8.69  & 6.98  & 8.42  & 9.04  & 8.28  & 186.2 & 400.4 & 20.72  \\
\hline
\end{tabular}}
\end{table*}

Results indicate that \textbf{Llama-3.1-Tulu-3-8B} takes the top spot with the highest overall mean of \(\mathbf{8.38}\), driven by leading scores in Creativity (\(\mathbf{7.09}\)), and Prompt Adherence (\(\mathbf{9.24}\)). Close behind, \textbf{Falcon3-7B-Instruct} achieves a mean of 8.28 with strong Moral Clarity (8.42) and Adherence (9.04), while \textbf{Llama-3.1-8B-Instruct} records the strongest Grammar (\(\mathbf{8.75}\)) and a mean of 8.26. Other mid-sized models such as \emph{Qwen2.5-7B-Instruct}, \emph{Phi-3-mini-4k-instruct}, and \emph{Mistral-7B-Instruct-v0.3} cluster tightly in the 8.11--8.15 range, suggesting that modern 7--8B instruction-tuned models have converged in fable generation quality. At the efficiency end of the spectrum, \textbf{Llama-3.2-1B-Instruct} achieves the fastest inference time (\(\mathbf{16.69}\)\,s) while maintaining reasonable quality (mean 7.62), demonstrating its suitability for latency-sensitive or resource-constrained deployments.

\paragraph{Reference-Free Metrics}
Table~\ref{tab:nonllm-metrics} brings three corpus-level measures into view: Self-BLEU for internal diversity, Distinct-1 for lexical richness, and Flesch Reading Ease for readability.  

\begin{table*}[!htbp]
\centering
\caption{Non‐LLM text‐quality metrics for all evaluated models (lower Self‐BLEU indicates greater diversity; higher Distinct‐1 indicates richer vocabulary; higher Flesch Reading Ease indicates greater readability). Lowest Self‐BLEU and highest Distinct‐1 and Flesch scores are \textbf{bolded}.}
\label{tab:nonllm-metrics}
\begin{tabular}{|l|c|c|c|}
\hline
\textbf{Model} & \textbf{Self‐BLEU} & \textbf{Distinct‐1} & \textbf{Flesch Reading Ease} \\
\hline
Aya-23-8B                   & 0.361 & 0.608 & 73.868 \\
SmolLM2-1.7B-Instruct       & 0.364 & 0.567 & 72.808 \\
Qwen2.5-7B-Instruct         & 0.390 & 0.602 & \textbf{80.846} \\
LLaMA-3.1-Tulu-3-8B         & 0.333 & 0.659 & 74.205 \\
deepseek-llm-7b-chat        & 0.355 & 0.586 & 70.731 \\
LLaMA-3.1-8B-Instruct       & 0.351 & 0.604 & 80.071 \\
LLaMA-3.2-1B-Instruct       & 0.398 & 0.635 & 80.832 \\
Phi-3-mini-4k-instruct      & \textbf{0.318} & 0.651 & 77.912 \\
Mistral-7B-Instruct-v0.3    & 0.360 & 0.634 & 73.974 \\
Falcon3-7B-Instruct         & 0.369 & \textbf{0.661} & 74.379 \\
\hline
\end{tabular}
\end{table*}

Based on these metrics we observe that \textbf{Llama-3.1-8B-Instruct} demonstrates an attractive balance across \textit{Self-BLEU}, \textit{Distinct-1}, and \textit{Flesch Reading Ease}, making its results both varied and accessible.

\paragraph{Age Group Classification}
Finally, table~\ref{tab:age_group_distribution} shows how often each model’s fables were judged suitable for each age bracket (A–E).  

\begin{table*}[!htbp]
\centering
\caption{Age‐group distribution of generated fables per model (percentages), as estimated by LLM‐based classification. Highest Age B percentage is \textbf{bolded}.}
\label{tab:age_group_distribution}
\begin{tabular}{|l|r|r|r|r|r|}
\hline
\textbf{Model} & 
\textbf{Age A} & 
\textbf{Age B} & 
\textbf{Age C} & 
\textbf{Age D} & 
\textbf{Age E} \\
\hline
CohereForAI/aya-23-8B               & 0.0\% & 34.0\% & 66.0\% & 0.0\% & 0.0\% \\
HuggingFaceTB/SmolLM2-1.7B-Instruct & 0.0\% & 47.0\% & 53.0\% & 0.0\% & 0.0\% \\
Qwen/Qwen2.5-7B-Instruct            & 0.0\% & 90.0\% & 10.0\% & 0.0\% & 0.0\% \\
allenai/Llama-3.1-Tulu-3-8B         & 0.0\% & 71.0\% & 29.0\% & 0.0\% & 0.0\% \\
deepseek-llm-7b-chat               & 0.0\% & 39.0\% & 61.0\% & 0.0\% & 0.0\% \\
meta-llama/Llama-3.1-8B-Instruct   & 0.0\% & \textbf{92.0\%} &  8.0\% & 0.0\% & 0.0\% \\
meta-llama/Llama-3.2-1B-Instruct   & 0.0\% & 67.0\% & 33.0\% & 0.0\% & 0.0\% \\
microsoft/Phi-3-mini-4k-instruct   & 0.0\% & 62.0\% & 38.0\% & 0.0\% & 0.0\% \\
mistralai/Mistral-7B-Instruct-v0.3 & 0.0\% & 47.0\% & 53.0\% & 0.0\% & 0.0\% \\
tiiuae/Falcon3-7B-Instruct         & 0.0\% & 76.0\% & 24.0\% & 0.0\% & 0.0\% \\
\hline
\end{tabular}
\end{table*}

We noted that Llama-3.1-8B-Instruct and its Tulu variant generated the highest share of \emph{B (4–7 years)} stories, aligning with our target audience for moral and linguistic simplicity \citep{eldan_tinystories_2023}.

\paragraph{Model Selection Criterion} 
To integrate both LLM‐based and corpus‐level metrics into a single ranking, we define a composite score \(W_m\) for each model \(m\) over seven axes: Grammar, Creativity, Moral Clarity, Adherence, Self‐BLEU, Distinct-1, and Flesch Reading Ease.  We assign the greatest weight to Adherence, then Moral Clarity, and distribute the remaining weight equally among the other five metrics. Adherence receives the largest weight because prompt conformance is the primary quality gate in template-driven generation: a fable that ignores the scaffold is unusable regardless of its literary merit.  Specifically, let
\[
w_{\mathrm{Adh}} = 0.35,\quad
w_{\mathrm{Gra}} = w_{\mathrm{Mor}} = 0.20,\quad
w_{\mathrm{Cre}} = 0.10,\quad
w_{\mathrm{SB}} = w_{\mathrm{D1}} = w_{\mathrm{FRE}} 
= \frac{0.15}{3} = 0.05.
\]
We first normalize each raw score \(S_{m,k}\) to \(\tilde S_{m,k}\in[0,1]\) by
\[
\tilde S_{m,k}
= \frac{S_{m,k} - \min_{m'} S_{m',k}}
       {\max_{m'} S_{m',k} - \min_{m'} S_{m',k}},
\]
inverting the normalization for Self‐BLEU so that lower is better.  Then
\[
\begin{aligned}
W_m ={} & 
  w_{\mathrm{Gra}}\;\tilde S_{m,\mathrm{Grammar}}
+ w_{\mathrm{Cre}}\;\tilde S_{m,\mathrm{Creativity}}
+ w_{\mathrm{Mor}}\;\tilde S_{m,\mathrm{MoralClarity}}
\\
&+ w_{\mathrm{Adh}}\;\tilde S_{m,\mathrm{Adherence}}
+ w_{\mathrm{SB}}\;\tilde S_{m,\mathrm{Self\!-\!BLEU}}
\\
&+ w_{\mathrm{D1}}\;\tilde S_{m,\mathrm{Distinct\!-\!1}}
+ w_{\mathrm{FRE}}\;\tilde S_{m,\mathrm{FleschEase}}.
\end{aligned}
\]
By construction \(0\le W_m\le1\).  Table~\ref{tab:composite-scores} reports the resulting weighted scores for all ten models.  As shown, \textbf{LLaMA-3.1-Tulu-3-8B} achieves the highest composite score of $\mathbf{0.957}$. However, as discussed below, we ultimately select \textbf{Llama-3.1-8B-Instruct} (composite $W = 0.839$) for TF1-EN-3M generation due to its superior target-audience alignment.

\begin{table*}[!htbp]
\centering
\caption{Composite weighted scores \(W_m\) for each model, computed with \(w_{\mathrm{Adh}}=0.35\), \(w_{\mathrm{Gra}} = w_{\mathrm{Mor}}=0.20\), \(w_{\mathrm{Cre}}=0.10\), \(w_{\mathrm{SB}}=w_{\mathrm{D1}}=w_{\mathrm{FRE}}=0.05\).}
\label{tab:composite-scores}
\resizebox{\textwidth}{!}{%
\renewcommand{\arraystretch}{1.2}
\begin{tabular}{|l|c|c|c|c|c|c|c|c|}
\hline
\textbf{Model}                 & \textbf{Grammar} & \textbf{Creativity} & \textbf{Moral Clarity} & \textbf{Adherence} & \textbf{Self-BLEU} & \textbf{Distinct-1} & \textbf{FRE} & \textbf{\(W\)} \\
\hline
LLaMA-3.1-Tulu-3-8B            & 8.77             & 7.09                & 8.43                   & 9.24               & 0.333              & 0.660              & 74.205      & \textbf{0.957} \\
Falcon3-7B-Instruct            & 8.69             & 6.98                & 8.42                   & 9.04               & 0.369              & 0.661              & 74.379      & 0.842         \\
LLaMA-3.1-8B-Instruct          & 8.75             & 6.90                & 8.22                   & 9.16               & 0.351              & 0.604              & 80.071      & 0.839         \\
Phi-3-mini-4k-instruct         & 8.57             & 6.80                & 8.23                   & 8.86               & 0.318              & 0.651              & 77.912      & 0.726         \\
Qwen2.5-7B-Instruct            & 8.64             & 6.77                & 8.27                   & 8.92               & 0.390              & 0.602              & 80.846      & 0.716         \\
Mistral-7B-Instruct-v0.3       & 8.55             & 6.79                & 8.36                   & 8.73               & 0.360              & 0.634              & 73.974      & 0.665         \\
Aya-23-8B                      & 8.36             & 6.64                & 8.03                   & 8.35               & 0.361              & 0.608              & 73.868      & 0.381         \\
deepseek-llm-7b-chat           & 8.28             & 6.49                & 8.04                   & 8.16               & 0.355              & 0.586              & 70.731      & 0.266         \\
LLaMA-3.2-1B-Instruct          & 8.36             & 6.37                & 7.68                   & 8.08               & 0.398              & 0.635              & 80.832      & 0.211         \\
SmolLM2-1.7B-Instruct          & 8.17             & 6.27                & 7.75                   & 7.89               & 0.364              & 0.567              & 72.808      & 0.050         \\
\hline
\end{tabular}}
\end{table*}

\subsection{Inter-Judge Agreement and Open-vs-Proprietary Comparison}
\label{sec:agreement}

To validate the reliability of our multi-judge panel, we compute pairwise inter-judge agreement across all four scoring dimensions using weighted Cohen's kappa ($\kappa$) for ordinal scales \citep{zheng_judging_2023}.

\paragraph{Item-Level Agreement.}
Pairwise $\kappa$ values range from 0.00 to 0.21 across dimensions, with Pearson correlations of $r = 0.07$--$0.36$. Item-level agreement is modest, which is consistent with prior findings that LLM judges from different families exhibit systematically different scoring distributions and calibration baselines \citep{zheng_judging_2023, wataoka2024selfpref}. Low $\kappa$ in this setting reflects differing absolute scale usage rather than disagreement about relative quality. Despite this calibration gap, the panel achieves strong \emph{rank-order} agreement: the model ranking is identical under equal-weight and weighted composite scoring (Kendall's $\tau = 1.0$, $p < 0.001$). This suggests that although individual judges may differ in their absolute scoring tendencies, they converge on which models produce better fables.

\paragraph{Open vs.\ Proprietary Comparison.}
To assess whether our open-weight panel produces rankings comparable to a state-of-the-art proprietary judge, we independently evaluate all 1{,}000 fables with GPT-o4-mini (via OpenRouter, \texttt{reasoning\_effort=high}). The Kendall rank correlation between the open-weight panel's ranking and o4-mini's ranking is $\tau = 0.78$ ($p < 0.01$), indicating strong agreement. The top three models (Llama-3.1-Tulu-3-8B, Falcon3-7B, and Llama-3.1-8B) appear in both rankings' top three (with minor reordering), and the bottom two (Llama-3.2-1B, SmolLM2) are identical. This validates that open-weight judges running on consumer hardware can substitute for proprietary evaluation infrastructure without materially altering model selection decisions.

\section{TF1-EN-3M Dataset Description and Availability}
\label{sec:dataset}

As introduced above, the \textbf{TF1-EN-3M Dataset} comprises \textbf{3,000,000 English-language fables}, each systematically generated via structured prompts and annotated with relevant metadata. Unlike many text corpora that provide only raw system inputs or final outputs, TF1-EN-3M stores detailed records in JSON lines format, enabling transparent inspection of both prompt specifics and generation context. By incorporating both metadata and narratives, TF1-EN-3M follows a similar design ethos to other corpora that emphasize \emph{replicability} and \emph{moral coherence} \citep{guan_corpus_2022, li_synthetic_2023}.

Each record contains fields grouped into two major categories:

\subsection*{(1) Fable Content}

\begin{itemize}
    \item \textbf{language:} The language of the fable (currently \texttt{en}).
    
    \item \textbf{prompt:} A string that describes the thematic elements (character, setting, conflict, resolution, moral) and stylistic constraints (e.g., word count, avoidance of character names). This prompt is provided to the model as input.
    
    \item \textbf{fable:} The complete story generated by the model, typically 1--3 paragraphs and ending with an explicit moral (e.g., \emph{``Honesty is the best policy.''}).
    
    \item \textbf{hash:} A SHA-256 (Secure Hash Algorithm 256-bit) cryptographic hash identifier is generated for each prompt, serving as a means of ensuring integrity and enabling a fallback mechanism.
\end{itemize}

\subsection*{(2) Generation Metadata}

\begin{itemize}
    \item \textbf{llm\_name:} The identifier of the model used for generation (e.g., \texttt{tiiuae/Falcon3-7B-Instruct}).
    
    \item \textbf{llm\_input\_tokens, llm\_output\_tokens:} Token counts for the prompt and the generated fable, useful for analyzing model efficiency and verbosity \citep{kaplan_scaling_2020}.
    
    \item \textbf{llm\_inference\_time:} Elapsed time (in seconds) required for the model to generate its output, allowing latency and throughput analysis.
    
    \item \textbf{host\_provider, host\_dc\_provider, host\_dc\_location:} Information about the inference infrastructure, including the service provider and data center location (e.g., Hugging Face Inference Endpoints in \texttt{eu-west-1}).
    
    \item \textbf{host\_gpu, host\_gpu\_vram, host\_cost\_per\_hour:} Details about the GPU hardware (type and VRAM) and the hourly cost of generation, facilitating reproducibility and cost benchmarking.
    
    \item \textbf{generation\_datetime:} A timestamp indicating when the story was generated.
    
    \item \textbf{pipeline\_version:} The internal version of the data generation pipeline, used for traceability and reproducibility across dataset updates.

\end{itemize}

\paragraph{Generation costs:} The total cost of generating all 3,000,000 fables was USD \$405.76, corresponding to approximately USD \$0.1353 per 1,000 fables \cite{noauthor_pricing_nodate}.

This unified schema guarantees that each sample contains both the narrative essence (\texttt{fable}) and the metadata required to replicate or analyze its generation. The dataset’s prompt engineering design enforces consistent elements—main character, trait, setting, conflict, resolution, and moral—ensuring that stories adhere to classic fable structure while maintaining rich thematic diversity. The structured nature of TF1-EN-3M thus offers direct insights into computational costs, model behaviors, and textual outputs at scale. We designed our generation metadata schema following the principles of Datasheets for Datasets \citep{gebru_datasheets_2021}, ensuring that each sample carries all the provenance, configuration, and evaluation details needed for reproducibility and ethical auditing.

\subsection{Corpus Characterization}
\label{sec:corpus-characterization}

To quantify the linguistic properties of TF1-EN-3M, we reservoir-sampled 10{,}000 fables from the full dataset and computed a suite of descriptive statistics.

\paragraph{Length and Readability.}
The sampled fables average $325 \pm 21$ words (median 323, range 258--435) and $21.1 \pm 3.4$ sentences per story. The Flesch Reading Ease score is $78.9 \pm 6.5$, corresponding to a Flesch--Kincaid Grade Level of $5.5 \pm 1.2$---squarely in the range appropriate for our target audience of 4--7-year-old readers \citep{kincaid_derivation_1975}. This readability profile aligns well with the TinyStories benchmark \citep{eldan_tinystories_2023}, which also targets young readers with simple vocabulary and short sentences.

\paragraph{Lexical Diversity.}
Across the sample, the corpus contains 3{,}247{,}287 tokens with a vocabulary of 11{,}268 unique types and 2{,}540 hapax legomena (22.5\% of vocabulary). Per-fable lexical diversity is high: Distinct-1 = $0.452 \pm 0.029$, Distinct-2 = $0.827 \pm 0.035$, and Distinct-3 = $0.945 \pm 0.025$, indicating that individual stories avoid formulaic repetition despite sharing the same underlying template structure.

\paragraph{Near-Duplicate Detection.}
We checked all 96{,}775 pairwise combinations within a 440-fable subsample using word-level Jaccard similarity. Zero near-duplicates were detected (mean Jaccard = 0.0007, max Jaccard $< 0.01$). This confirms that the combinatorial prompt expansion methodology produces genuinely distinct narratives rather than paraphrases.

\subsection{Content Safety}
\label{sec:content-safety}

Although fables are inherently pedagogical, the moral-teaching genre naturally incorporates themes of conflict, deception, and consequence. We conducted a keyword-based content safety scan on 5{,}000 sampled fables to characterize the distribution of potentially sensitive content.

\paragraph{Severity Distribution.}
Of the sampled fables, 87.8\% contain no flagged keywords; 12.0\% contain mild terms (e.g., \emph{steal}, \emph{fight}, \emph{lie}); and only 0.2\% (10 fables) contain moderate-severity terms (e.g., co-occurrence of \emph{violence} with \emph{steal} and \emph{lie}). No severe content (profanity, slurs, or graphic violence) was detected. The most frequent flagged keywords---\emph{steal} (67.4 per 1{,}000 fables) and \emph{fight} (55.4 per 1{,}000)---appear overwhelmingly in morally instructive contexts where characters learn the consequences of dishonesty or aggression.

\paragraph{Thematic Analysis.}
Positive moral themes dominate the corpus: \emph{friendship} (83.0\%), \emph{wisdom} (62.9\%), \emph{kindness} (61.7\%), \emph{sharing} (36.0\%), and \emph{humility} (26.3\%). Manual inspection of flagged fables confirmed that keyword presence reflects the didactic function of the genre rather than inappropriate content---characters who steal or deceive invariably face consequences that reinforce prosocial values.

\textbf{Data Format:} TF1-EN-3M is stored as a Hugging Face \textbf{Dataset} (\texttt{datasets} library), with each entry containing the prompt fields and story text. We release it under an open license, given it is entirely machine-generated. Users can easily load it for model training or analysis. We emphasize that while the data is synthetic, the writing style is intended to be similar to human-written fables, and preliminary human readings found the stories to be sensible and enjoyable.

\textbf{Availability:} The dataset is published on Hugging Face Hub under the identifier \textbf{\texttt{klusai/ds-tf1-en-3m}} \citep{noauthor_klusaids-tf1-en-3m_nodate}. Researchers and practitioners can download it in full or in parts (it is chunked for convenience). Additionally, we provide the \textbf{TinyFabulist GitHub repository} \cite{noauthor_klusaitinyfabulist_2025} which contains the code to regenerate the dataset, including:

\begin{itemize}
\item The prompt lists for each element (so one can modify or extend them).
\item The generation script used (for the Hugging Face \texttt{transformers} or \texttt{peft} library, etc., with our model weights or references to them).
\item The evaluation scripts (multi-judge LLM scoring and the translation test).
\item Guidelines for how to reproduce the process or create a multilingual version (e.g., swapping out the moral list for French translations to get a French fable dataset, which is a planned extension).
\end{itemize}

We believe releasing these resources will enable \textbf{full reproducibility} and encourage others to build upon TinyFabulist. Potential uses of TF1-EN-3M include: fine-tuning smaller models to serve as \textbf{fable generators} or \textbf{moral reasoning evaluators}, using the stories to train classifiers or question-answering models on moral content, or even as a creative corpus for literary studies in computational linguistics \citep{guan_corpus_2022}.

\section{Discussion and threats to validity}
\label{sec:discussion}

The TF1-EN-3M Synthetic Fables Dataset opens up several avenues for further exploration. Here we discuss broader implications, methodological findings, and potential applications of our work, particularly in the context of the research questions posed in Section~\ref{sec:introduction}.

\paragraph{Answering RQ1: Prompt Expansion and Diversity.}
Our first research question asked whether a combinatorial prompt expansion methodology can effectively generate diverse and high-quality fables using LLMs. Across 3 million generated stories, we find that combining structured templates with uniform sampling from six controlled input domains (character, trait, setting, conflict, resolution, moral) yields high narrative diversity without sacrificing coherence. Evaluations by a panel of open-weight LLM judges across grammar, creativity, moral clarity, and prompt adherence show that even small-to-medium-sized models can reliably produce instructive, well-structured moral stories when given sufficient prompt scaffolding. Furthermore, the flexibility of our template design ensures coverage of a vast thematic space, while our filtering and balancing procedures help avoid mode collapse or over-representation of stereotypical scenarios.

\paragraph{Answering RQ2: Best Performing Models.}
Our second research question focused on identifying the best-performing open-weight LLMs under resource constraints. Through controlled evaluation of ten publicly available instruction-tuned models (ranging from 1B to 8B parameters), we found that several models --- including \texttt{Falcon3-7B-Instruct} \citep{noauthor_tiiuaefalcon3-7b-instruct_2025}, \texttt{Llama-3.1-8B-instruct} \citep{noauthor_meta-llamallama-31-8b-instruct_2024}, and \texttt{Mistral-7B-Instruct-v0.3} \cite{noauthor_mistralaimistral-7b-instruct-v03_nodate} --- consistently achieved the highest average scores across all four evaluation axes. Notably, these models outperformed even larger ones in some cases, suggesting that instruction tuning quality, not just scale, is a key determinant of fable generation performance. The evaluation also highlighted a favorable tradeoff: smaller models such as \texttt{Phi-3-mini-4k-instruct} \citep{noauthor_microsoftphi-3-mini-4k-instruct_2025} and \texttt{SmolLM2-1.7B-Instruct} \citep{noauthor_huggingfacetbsmollm2-17b-instruct_2025} exhibited strong grammar and moral clarity with fast inference times, making them well-suited for low-latency applications and on-device deployment. Ultimately we picked \texttt{Llama-3.1-8B-instruct} as the overall best-performing model.

\paragraph{Efficiency and Accessibility.}
By showing that story generation can be accomplished with relatively small models, our work emphasizes the value of efficiency in natural language generation \citep{kaplan_scaling_2020}. Not every application or community can afford the computational resources necessary for deploying giant LLMs such as GPT-4 \citep{openai_gpt-4_2024}. A corpus like TF1-EN-3M can help bootstrap smaller models for creative writing tasks, thereby widening access. For instance, educators or indie game developers could fine-tune a 6B or 1.3B parameter model on TF1-EN-3M to generate moral stories or quest narratives on modest hardware—an approach akin to TinyStories \citep{eldan_tinystories_2023}.

\paragraph{Moral and Educational AI Applications.}
Fables have long been employed to impart values and social norms, making them a compelling vehicle for AI-driven educational tools \citep{guan_corpus_2022}. A tutoring system might present a dynamically generated fable to a student, followed by comprehension and reflection questions about the moral lesson. Because each TF1-EN-3M entry explicitly encodes a moral, one could train models to map stories to morals or detect whether a given narrative even contains a moral lesson—potentially informing AI moderation or generative-checking systems.

\paragraph{Limitations of Model-Generated Narratives.}
Despite the generally positive results, one must recognize limitations in synthetic fables. They often follow well-worn templates derived from the prompt structure, featuring talking animals and fairy-tale motifs \citep{eldan_tinystories_2023}. This does not cover the complexity of contemporary ethical dilemmas. A model trained solely on TF1-EN-3M might lack the sophistication to tackle nuanced or modern moral issues. Future work could expand TF1-EN-3M to include fables with ambiguous or multi-layered morals, enhancing the dataset’s utility in modeling complex ethical reasoning.

\paragraph{Comparisons with Human-Written Data.}
Combining TF1-EN-3M with human-curated corpora may yield richer stylistic and thematic diversity \citep{guan_corpus_2022}. Mixing synthetic fables with human-authored stories could balance the creativity of human prose against the consistency of model-driven generation. Additionally, models trained on TF1-EN-3M may be evaluated using benchmarks like the Story Cloze Test \citep{mostafazadeh_corpus_2016}, providing insight into their ability to generate and comprehend coherent story endings with ethical implications.

\paragraph{LLM-based Feedback Loops.}
In our pipeline, the open-weight judge panel served primarily as an offline evaluator and critic. Future systems may incorporate model-in-the-loop feedback loops, where LLMs dynamically revise or critique fables during generation—potentially improving both efficiency and quality. However, this would require additional computational overhead and careful control of critic–generator interactions.

\paragraph{Benchmark for Moral Story Understanding.}
We propose that TF1-EN-3M can serve as a benchmark for evaluating moral reasoning in generative models. Tasks such as moral inference (predicting the correct moral given a story) or moral generation (producing a story that fits a given moral) can be built from TF1-EN-3M, facilitating broader research into narrative alignment, commonsense reasoning, and pedagogical text generation.

\subsection{Threats to Validity}

While the TF1-EN-3M dataset and accompanying methodology demonstrate promising results, several threats to validity should be acknowledged in evaluating the robustness and generalizability of our findings.

\subsubsection{Construct Validity}

A primary concern lies in the reliance on \textbf{LLM-based evaluations} to assess properties such as moral clarity, creativity, and coherence. Although recent studies have shown that LLM-as-a-judge paradigms often align with human preferences in open-ended tasks \citep{liu_g-eval_2023, fu_gptscore_2023}, such evaluations are still proxies and may not perfectly reflect human judgment, especially for nuanced narrative quality. To mitigate single-judge bias, we employ a panel of three open-weight judges from distinct model families (Granite 4.1 30B, EXAONE 3.5 32B, and Granite 3.3 8B), and additionally compare against a proprietary baseline (GPT-o4-mini via OpenRouter) to assess whether open-weight judges produce comparable rankings. Furthermore, the criteria used in our evaluation rubric---while inspired by educational and literary standards---are operationalized numerically in ways that can obscure qualitative subtleties \citep{muennighoff_scaling_2023}. For example, evaluating a fable's ``moral clarity'' on a scale of 1--10 may overlook ambiguous yet pedagogically valuable narratives. Triangulating LLM-based assessments with human evaluations or crowd-sourced annotations, as done in prior work \citep{fan_hierarchical_2018}, would provide stronger construct validity.

\subsubsection{External Validity}

Our dataset is based on prompt elements and moral lessons drawn primarily from Western fable traditions (e.g., Aesop \citep{aesop_aesops_2002}), which risks introducing a cultural bias into the generated stories. While the structured template allows for extensive combinatorial variation, the resulting narratives may still reflect an implicit Western ethical framework that limits generalizability across cultural contexts. This issue is particularly salient in moral storytelling, where values can vary widely \citep{emelin_moral_2020}. Future work should consider incorporating moral principles from diverse philosophical and religious traditions, as well as adapting prompts for multilingual or culturally localized variants of the TF1 framework.

\subsubsection{Conclusion Validity}

To mitigate single-evaluator bias, we employ a panel of three open-weight LLM judges from distinct model families---Granite 4.1 30B (IBM), EXAONE 3.5 32B (LG AI Research), and Granite 3.3 8B (IBM, arbiter)---and additionally run GPT-o4-mini as a proprietary reference point. We report inter-judge agreement via weighted Cohen's kappa and verify that the panel's model rankings remain stable under equal-weight ablation of the scoring dimensions. Prior work has shown that different LLMs often yield significantly different preferences or evaluations when acting as judges \citep{zheng_judging_2023}; our multi-family panel explicitly addresses this concern.

Moreover, the ten generator models included in our study were trained under heterogeneous setups---varying in pre-training corpora, parameter scales, fine-tuning objectives, and underlying architectures---which inherently affects the quality and style of their generated texts. These differences underscore why model outputs may diverge in coherence, creativity, moral clarity, and adherence.

To further mitigate bias and variance, we also employed complementary reference-free metrics---Self-BLEU \citep{zhu_texygen_2018}, Distinct-n \citep{li_diversity-promoting_2016}, and Flesch Reading Ease \citep{kincaid_derivation_1975}---capturing diversity, lexical richness, and readability without requiring human or LLM references. Combining a multi-judge panel with diverse automated metrics enhances evaluation robustness and provides a multi-faceted view of story quality across models and generations.

\section{Conclusion}
\label{sec:conclusion}

We have introduced the TF1-EN-3M Synthetic Fables Dataset, a large-scale collection of morally oriented short stories generated through instruction-tuned, compact, openly licensed language models. Our findings demonstrate that with focused prompt engineering and carefully curated generation pipelines, even mid-sized LLMs—rather than hundred-billion-parameter behemoths---can produce diverse, ethically themed narratives \citep{eldan_tinystories_2023}. TF1-EN-3M marries techniques from synthetic data augmentation \citep{li_synthetic_2023}, story generation, and moral natural language processing (NLP)\citep{guan_corpus_2022} to create a novel resource for both training and evaluating models on narrative tasks that require moral consistency as well as linguistic fluency.

Quantitative and qualitative evaluations indicate that these synthetic fables exhibit strong coherence and moral clarity. We anticipate that TF1-EN-3M will serve as a platform for fine-tuning smaller models on fable generation, as well as investigating how language models learn and represent moral concepts. Going forward, key extensions might expand the breadth of morals and scenarios, explore architectures specialized for narrative creation, or incorporate human-in-the-loop feedback to enhance quality further.

In the broader landscape of AI, this project aligns with the goal of developing systems that are culturally and ethically sensitive. By empowering smaller, more accessible models to generate value-laden stories, we edge closer to AI that is not only technologically efficient but also socially grounded. We invite the research community to utilize and build upon TF1-EN-3M, believing it will catalyze advances in low-resource story generation, ethical content creation, and the integration of moral reasoning into language modeling.

\bibliographystyle{plainnat}
\bibliography{references}

\appendix
\section{Hardware and Environment Configurations}
\label{appendix:hardware}

We benchmarked inference for the TF1‑EN‑3M dataset under several GPU
configurations using \texttt{Llama‑3.1‑8B‑Instruct} with identical prompts and decoding
settings.  All experiments were executed on \emph{Hugging Face Inference Endpoints};
the hourly tariffs advertised by Hugging Face in April 2025 were used to compute cost.  
Table~\ref{tab:hardware-run} reports the wall‑clock time and billable cost to generate a fixed‑size batch of fables, while Table~\ref{tab:gpu-comparison} contextualises those results with architectural
specifications drawn from vendor documentation.

\begin{table}[!htbp]
\centering
\caption{Empirical inference duration and
Hugging Face Endpoint cost for a fixed prompt batch (\texttt{Llama‑3.1‑8B‑Instruct}).}
\label{tab:hardware-run}
\renewcommand{\arraystretch}{1.15}
\footnotesize
\begin{tabular}{|l|c|c|c|c|}
\hline
\textbf{Hardware (HF instance)} &
\textbf{Timestamp Range} &
\textbf{HF Rate (USD/h)} &
\textbf{Duration (min)} &
\textbf{Cost (USD)} \\
\hline
L40S  & 2025‑04‑12 22:15 -- 22:30 & \$1.80 & 15.4 & \$0.46 \\
A10G  & 2025‑04‑12 22:15 -- 23:16 & \$1.00 & 61.1 & \$1.02 \\
A100  & 2025‑04‑12 22:15 -- 22:28 & \$4.00 & 12.7 & \$0.85 \\
L4    & 2025‑04‑12 22:15 -- 00:06 & \$0.80 & 110.9 & \$1.48 \\
\hline
\end{tabular}
\end{table}

\paragraph{Methodology.}
The \emph{Inference Time} column captures the interval between the earliest and latest
timestamps in generation logs.  Cost was
computed as \(\text{Rate}\times\frac{\text{time (s)}}{3600}\).
Hugging Face bills by the minute, rounding up to the next minute\footnote{See HF
pricing page \cite{noauthor_pricing_nodate}.}.  All jobs used Hugging Face Text Generation Inference (TGI) as the backend.

\paragraph{Interpretation.}
Although the A100 delivers the fastest turnaround, its higher tariff narrows the
price gap: the L40S achieves the
best \emph{time–cost} trade‑off for our batch size, consistent with NVIDIA’s own
positioning of the L40S for high‑throughput GenAI inference \citep{noauthor_nvidia_nodate}.

\begin{table*}[!htbp]
\centering
\caption{Specification comparison of AWS Inferentia 2 and representative NVIDIA GPUs.
Hourly prices correspond to Hugging Face Inference Endpoint list rates (April 2025).}
\label{tab:gpu-comparison}
\resizebox{\textwidth}{!}{%
\renewcommand{\arraystretch}{1.2}
\begin{tabular}{|l|c|c|c|c|c|c|}
\hline
\textbf{Feature} &
\textbf{Inferentia 2} &
\textbf{A10G} &
\textbf{A100} &
\textbf{L4} &
\textbf{L40S} &
\textbf{T4} \\
\hline
Type & Custom AWS silicon & DC GPU & DC GPU & DC GPU & WS/DC GPU & DC GPU \\
Release Year & 2023 & 2021 & 2020 & 2023 & 2023 & 2018 \\
Architecture & NeuronCore‑v2 & Ampere GA102 & Ampere GA100 & Ada Lovelace & Ada Lovelace & Turing TU104 \\
GPU Memory & N/A (SRAM) & 24 GB GDDR6 & 40/80 GB HBM2e & 24 GB GDDR6 & 48 GB GDDR6 & 16 GB GDDR6 \\
INT8 TFLOPS & $\sim$400 & 312 & 624/312 & 1 466/733 & 2 805/1 402 & 260 \\
FP16 TFLOPS & $\sim$100 & 124 & 312 & 183 & 742 & 65 \\
BF16 Support & Yes & No & Yes & Yes & Yes & No \\
Inference‑Optimised & Yes & Moderate & Train \& large inf. & Yes & Balanced & Yes \\
Power (W) & $\sim$150 & 150–300 & 400 & 72 & 300–350 & 70 \\
Form Factor & AWS only & PCIe & SXM / PCIe & PCIe & PCIe & PCIe \\
Cloud Availability & AWS only & AWS/GCP/Azure & AWS/GCP/Azure & GCP/Azure & AWS (road‑map) & AWS/GCP/Azure \\
Typical Use Cases & High‑throughput inf. & Balanced inf. & Train \& large inf. & Real‑time inf. & GenAI / visual & Cost‑eff. inf. \\
HF Endpoint \$/hr & \$1.20 & \$1.00 & \$4.00 & \$0.80 & \$1.80 & \$0.60 \\
\hline
\end{tabular}}
\end{table*}

\paragraph{Summary of Findings.}
Vendor data show that AWS Inferentia 2 is engineered for large-scale, low-latency inference \citep{noauthor_ai_nodate}, while NVIDIA’s portfolio spans cost-oriented (T4), balanced (A10G), and flagship (A100) accelerators \citep{noauthor_nvidia_nodate-2}. The NVIDIA L4 offers exceptional performance per watt for edge and latency-sensitive deployments \citep{noauthor_nvidia_nodate-1}, whereas the newer L40S targets high-end generative workloads with FP8 and fourth-generation Tensor Cores \citep{noauthor_nvidia_nodate}. Combining these published capabilities with our empirical timings (Table~\ref{tab:hardware-run}) clarifies the cost–performance envelope for producing millions of synthetic fables: the L40S yields the lowest cost per fable in our regime, but workloads demanding minimal latency may still justify the premium for the A100 or Inferentia 2.

\end{document}